\documentclass[letterpaper, 10 pt, conference]{ieeeconf}  
\IEEEoverridecommandlockouts                         
\usepackage{graphics}
\usepackage{epsfig}
\usepackage{mathptmx}
\usepackage{amsmath}
\usepackage{amssymb}
\usepackage{hyperref}
\usepackage{cleveref}
\usepackage{cite}
\usepackage{booktabs}

\crefname{figure}{Fig.}{Figs.}
\Crefname{figure}{Fig.}{Figs.}

\crefname{table}{Table}{Tables}
\Crefname{table}{Table}{Tables}

\crefname{section}{Section}{Sections}
\Crefname{section}{Section}{Sections}

\crefname{subsection}{Section}{Sections}
\Crefname{subsection}{Section}{Sections}

\crefname{subsubsection}{Section}{Sections}
\Crefname{subsubsection}{Section}{Sections}

\crefname{equation}{Eq.}{Eqs.}
\Crefname{equation}{Eq.}{Eqs.}

\crefname{appendix}{Appendix}{Appendices}
\Crefname{appendix}{Appendix}{Appendices}

\title{\LARGE \bf
PneuTac: Tactile Manipulation with Soft Pneumatic Robots via Unified MPM-Gaussian Splatting Simulation
}

\author{Shaohong Zhong$^{1}$, Marco Pontin$^{1}$, Joe Watson$^{1}$, Perla Maiolino$^{1}$, and Ingmar Posner$^{1}$%
\thanks{$^{1}$Oxford Robotics Institute, Department of Engineering Science, University of Oxford, OX1 3PJ Oxford, U.K. 
        {\tt\small shaohong@robots.ox.ac.uk}}}%

\begin{document}

\maketitle
\thispagestyle{empty}
\pagestyle{empty}

\begin{abstract}
Soft robots and tactile sensors have demonstrated great potential in delicate manipulation tasks. Soft pneumatic robots enable safe contact through compliance, and vision-based tactile sensors offer high-resolution touch perception. However, learning tactile manipulation with compliant robots has been challenging, bottlenecked by the lack of efficient simulation. Existing simulators typically model them in isolation, and exhibit large calibration gaps that are difficult to overcome efficiently. We present PneuTac, a unified framework for tactile-feedback manipulation with soft pneumatic robots. We leverage the material point method (MPM) for modelling the dynamics of the soft robot and the deformable tactile membrane, and 3D Gaussian splatting (3DGS) for rendering. Real-to-sim modelling is done with a simple vision-based method, to then train action and perception networks for efficient simulation with surrogate models. We use the framework to drive a tactile-guided pipeline to collect demonstrations in simulation. Through experiments on a custom-designed pneumatic soft finger with a tactile sensing tip, together with additional cross-device evaluations, we show that PneuTac is capable of accurately modelling soft robots with tactile sensors, and that policies trained with simulation-augmented demonstrations outperform baselines trained on the same real data on three real-world contact-rich compliant manipulation tasks, making it a practical framework for tactile manipulation on compliant hardware.
\end{abstract}
\section{Introduction}
\label{sec:intro}

Soft robots and tactile sensors represent a natural pairing for delicate, contact-rich manipulation. Compliant pneumatic robots deform under contact and tolerate uncertainties during interactions~\cite{rus2015design}, whilst vision-based tactile sensors are highly sensitive to contact, using a camera behind a deformable gel to record high-resolution contact images that capture the gel indentation~\cite{yuan2017gelsight}. Recent developments in tactile-sensing-equipped soft robots also show great promise in utilising their synergies for tactile manipulation with compliant robots in delicate robotics tasks ~\cite{zhang2025pneugelsight}. 

However, harnessing the potential of this pairing has proven difficult. Current learning-based paradigms typically require a substantial amount of demonstration data, especially for tasks involving deformable materials and contact-rich interactions~\cite{duriez2013sofa,si2022taxim,si2024difftactile}. Data collection is slow and expensive on real soft hardware and accelerates wear and tear on the compliant components~\cite{zhong2025tactgen}. Simulation offers a natural solution, but the combination of a soft body and a vision-based tactile sensor with a soft gel is challenging for existing tools, which primarily consider them in isolation. Vision-based tactile simulators typically model the sensor in a rigid setting~\cite{si2022taxim,si2024difftactile,sun2024tacchi,you2026dot,li2025taccel}. On the other hand, soft robot simulations do not typically consider contact due to the extra modelling and computational complexities on top of the challenging non-linear dynamics of the robots themselves~\cite{dubied2022sors, zhong2025end}. Both also require significant amounts of data for real-to-sim modelling, and often need repeated calibration due to material or device degradation~\cite{duriez2013sofa, zhong2023touch,zhong2025tactgen,zhong2025end}. To our knowledge, no prior work has been able to jointly model compliant actuation and compliant tactile sensing for policy learning. 

\begin{figure*}[!tbp]
  \centering
  {\includegraphics[width=0.9\textwidth]{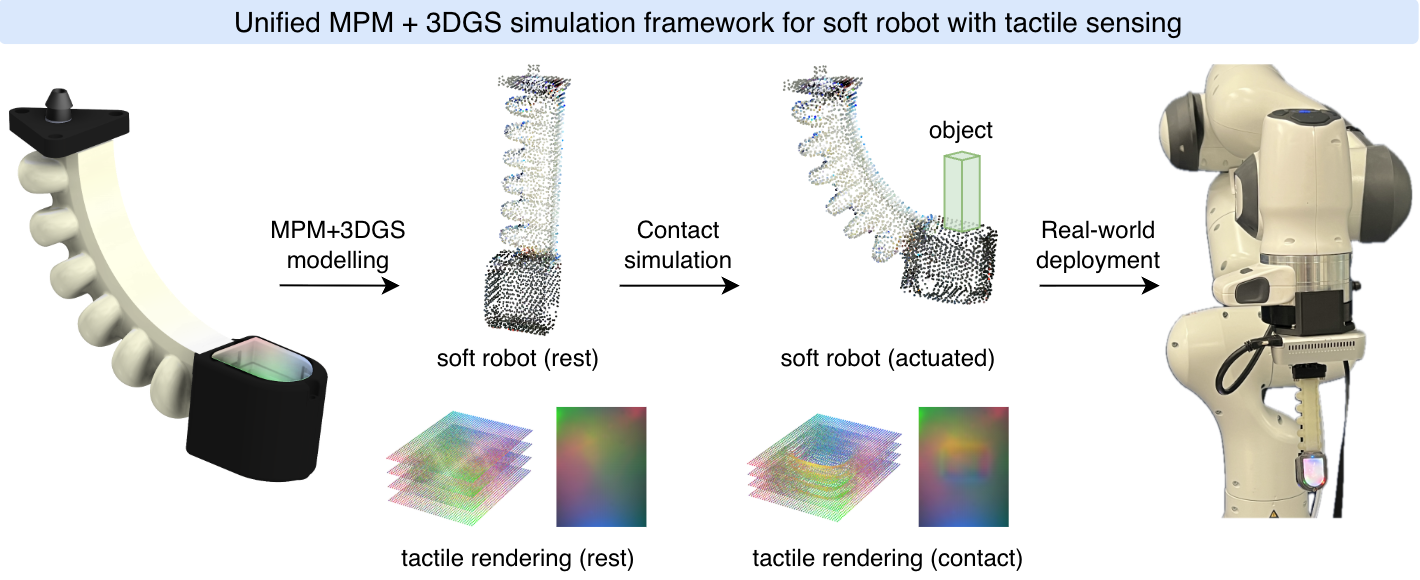}
  \caption{Overview of PneuTac. The framework leverages MPM-3DGS coupling for efficient real-to-sim modelling from CAD models. This supports accurate modelling of soft robot deformation and tactile contact rendering, which facilitates policy learning for real-world contact-rich compliant manipulation.}
  \label{fig:exp:overview}}
\vspace{-3mm}
\end{figure*}

We present PneuTac, a unified framework for tactile-based manipulation on soft pneumatic robots. We leverage the material point method (MPM) for dynamics modelling, and 3D Gaussian Splatting (3DGS) for the appearance model~\cite{sulsky1994particle, kerbl20233dgs}. GPU-based MPM simulations enable the fast simulation of the complex dynamics, whereas 3DGS models enable efficient rendering and real-to-sim modelling of the tactile sensor and the soft robot with simple vision-based methods. The framework also enables sim-to-real manipulation learning, where we leverage the framework in a tactile-guided pipeline for demonstration augmentation. We then train tactile-based manipulation policies and demonstrate transfer to real hardware with a small amount of fine-tuning, substantially increasing the policy performance compared to using the same amount of real-world data for training. An overview of the framework is shown in ~\cref{fig:exp:overview}.

This paper makes the following contributions.
\begin{enumerate}
\item A unified MPM-3DGS simulator that jointly simulates soft pneumatic actuators and vision-based tactile sensors with a deformable gel.
\item Data-efficient real-to-sim modelling for the devices through the MPM-3DGS representation, with cross-device validation on distinct pneumatic robot morphologies and tactile sensor instances.
\item Tactile-guided pipeline for simulation-based demonstration augmentation and sim-to-real validation on three tactile manipulation tasks on compliant hardware.
\end{enumerate}

\section{Related Works}
\label{sec:related}

\subsection{Vision-based tactile simulation}
Prior attempts at simulating vision-based tactile sensors have included example-based renderers such as Taxim~\cite{si2022taxim} and physics-based simulators that model the sensing elastomer as a deformable body. DiffTactile~\cite{si2024difftactile} uses a finite element model for the membrane and a learned optical module for the tactile image. Tacchi and Tacchi~2.0~\cite{chen2023tacchi,sun2024tacchi} use a moving least squares material point method for the elastomer paired with conventional rendering. A concurrent work, DOT-Sim~\cite{you2026dot}, calibrates an MPM membrane model against finite element ground truth from roughly twenty demonstration videos, and Taccel~\cite{li2025taccel} scales tactile simulation to parallel environments using incremental potential contact. In each case, the sensor simulations are mostly limited to coupling with rigid robot and object physics, and do not support attachment to compliant robots. Further, these approaches typically require a significant amount of data for rendering calibration. In contrast, PneuTac differs by enabling the tactile simulation on a soft robot, and by learning the tactile rendering from a single indentation image.

\subsection{Soft robot simulation}
Previous work on soft robot simulation has focused on numerical methods such as finite element methods~\cite{duriez2013sofa, zhong2025end} and GPU-accelerated multi-physics engines~\cite{genesis2024}. These approaches typically require significant amounts of effort for calibration to overcome the sim-to-real gap~\cite{dubied2022sors,gao2024residual,menager2025diff}. Additionally, there have been limited attempts to enable tactile sensing on these compliant robots~\cite{zhang2025pneugelsight,nguyen2026eletac}. These focus on robot construction and sensing, and do not support simulation for manipulation policy learning. PneuTac extends this line of work to learning for tactile-based manipulation with an efficient unified simulator.

\subsection{Physics-integrated Gaussian rendering}
A growing body of work integrates physics with 3D Gaussian Splatting~\cite{kerbl20233dgs}. PhysGaussian~\cite{xie2024physgaussian} drives Gaussian kernels with a material point method so that simulation and rendering share a single representation. PhysTwin~\cite{jiang2025phystwin} reconstructs interactive digital twins of deformable objects such as ropes and cloth from sparse video. Similar to PhysGaussian and PhysTwin, our approach leverages an MPM-3DGS pipeline for modelling deformable objects. However, instead of passive deformables, PneuTac applies it to actuated robots with tactile sensing for sim-to-real policy learning, which is not achieved by prior works.
\section{Methodology}
\label{sec:method}

\begin{figure*}[h]
  \centering
  {\includegraphics[width=0.95\textwidth]{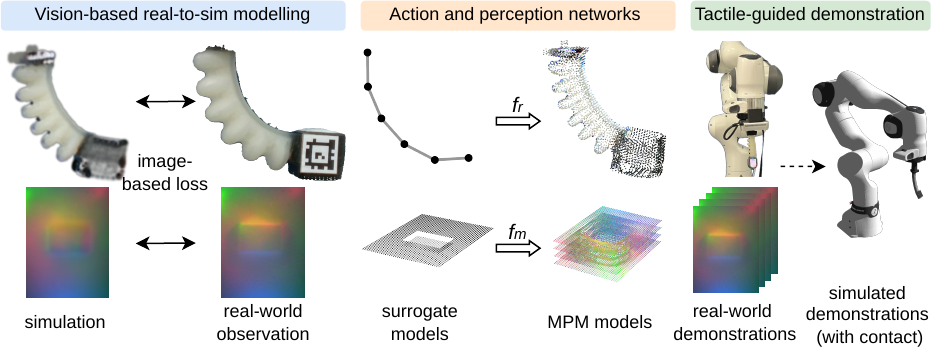}
  \caption{PneuTac pipeline. Left: image-based real-to-sim calibration. Middle: residual-torque prediction ($f_r$) and probe-to-MPM indentation mapping ($f_m$) for fast surrogate simulation. Right: generation of simulated demonstrations guided by real trajectories and tactile observations.}
  \label{fig:exp:method}}
  \vspace{-3mm}
\end{figure*}

The PneuTac framework has four components. 1) An MPM-3DGS simulator for modelling the dynamics and appearances of the soft robot and the tactile sensor. 2) A simple vision-based procedure for calibration of both devices. 3) Action and perception networks for efficient surrogate modelling. 4) A tactile-guided pipeline for using the calibrated simulator to collect demonstrations from a small set of real-world demonstrations. An overview is shown in~\cref{fig:exp:method}.

\subsection{MPM-3DGS Simulation}
\label{sec:method:sim}
We represent both the soft robot and the deformable gel of the vision-based tactile sensor using the material point method (MPM). MPM is a hybrid particle–grid scheme that tracks a deformable body as material particles while resolving its dynamics on a background grid, thus handling large deformation and contact robustly~\cite{sulsky1994particle}. To initialise the particle model, we leverage typically available CAD models for the soft robot and the tactile sensor, and sample particles from their mesh models. We additionally approximate the pneumatic bending chamber with a rigid multi-joint spine embedded along the finger's neutral axis using the Compatible Particle-in-Cell (CPIC) model to drive its motion~\cite{rus2015design,hu2018moving}. For simulating the behaviour of the deformable materials, we leverage an off-the-shelf implementation of the Neo-Hookean constitutive model~\cite{genesis2024}.

For learning the appearance model, inspired by prior work \cite{xie2024physgaussian,jiang2025phystwin}, we use the 3DGS pipeline coupled with the MPM model. 3D Gaussian Splatting represents a scene as a set of anisotropic 3D Gaussians, each carrying a position, covariance, colour, and opacity, which are rasterised into images and enable fast rendering. We leverage sampled particles from the MPM model to define the positions of 3D Gaussians and fix their scales, and only train the colour and opacities using \textit{one} cropped image of the physical devices. The motion of the particles in MPM thus results in coupled movements of the Gaussian means, which rasterise into realistic renderings of the robot's deformation. 

For the tactile sensor, in addition to the external appearance modelled jointly with the soft robot, we focus on accurately simulating the tactile image readings, which capture the indentation of the gel under internal LEDs within the sensor chamber. We leverage the mesh model of the gel to sample an additional set of MPM particles with coupled 3D Gaussians, and optimise the Gaussians using \textit{one} rest background image of the sensor. When the gel deforms under contact, the corresponding 3D Gaussians move and rasterise to a new tactile image with indentation. On top of that, we fit a slope- and depth-dependent contact photometric model for accurate modelling of the light effects from the gel indentation, explained in detail in \cref{sec:method:calib}.

\subsection{Real-to-Sim Modelling}
\label{sec:method:calib}
The MPM-3DGS simulation provides a realistic appearance model coupled with a dynamics model, thus allowing real-to-sim modelling using a simple vision-based procedure instead of the typically complicated setup required for system identification. For calibration of the soft robot model, we aim to identify the material parameters $\boldsymbol{\theta}_m$ for MPM simulation, as well as the actuation parameters $\boldsymbol{\theta}_a$ for mapping input pneumatic pressures $p$ to joint torques $\boldsymbol{\tau}_\mathrm{mpm}$, where $\boldsymbol{\theta}=(\boldsymbol{\theta}_m,\boldsymbol{\theta}_a)$. We calibrate the parameters in a quasi-static setting, where we actuate the robot with a fixed set of pressures and capture the deformation at equilibrium with a camera. Parameter identification is then performed using a sampling-based method with an image-based loss. 

For calibration of the tactile sensor reading, once the base 3DGS model is trained on the rest background image $\mathbf{I}_\mathrm{rest}$, its parameters are fixed, where each MPM particle is represented one-to-one by a 3D Gaussian with base colour $\mathbf{c}^0_g$. We then leverage a single indentation image $\mathbf{I}_\mathrm{indent}$ to calibrate a photometric model for contact. Assuming indentation on the top surface of the simulated gel MPM model, the top-layer per-particle vertical compression is interpolated and smoothed into an image-plane depth map $d$ with in-plane gradient $\nabla d=(g_x,g_y)$. The model adds to each Gaussian a colour increment linear in surface slope and depth, mimicking the sensor's directional coloured LEDs and depth-dependent contrast, $\mathbf{c}_g=\mathbf{c}^0_g + w_g[\mathbf{L}\nabla d_g + \mathbf{c}_d\bar{d}_g]$, where $\mathbf{L}=[\boldsymbol{\ell}x,\boldsymbol{\ell}y]\in\mathbb{R}^{3\times2}$ stacks the per-channel directional-illumination coefficients, $\nabla d_g$ is the surface slope at Gaussian $g$, $\bar{d}_g$ is the normalised indentation depth, $\mathbf{c}_d\in\mathbb{R}^3$ is the depth-contrast coefficient, and $w_g$ is a surface weight. Denoting the render by $\hat{\mathbf{I}}$, the coefficients of the photometric contact model $\{\boldsymbol{\ell}_x,\boldsymbol{\ell}_y,\mathbf{c}_d\}$ are estimated from a single registered pair $(\mathbf{I}_\mathrm{rest},\mathbf{I}_\mathrm{indent})$ and the matching MPM deformation $\boldsymbol{\Delta}_\mathrm{mpm}$ by minimising the photometric loss $\lVert\hat{\mathbf{I}}-\mathbf{I}_\mathrm{indent}\rVert_2^2$.

\subsection{Surrogate Models with Action and Perception Networks}
\label{sec:method:network}
For efficient simulation, instead of running the full MPM simulation, we train two separate networks to map the behaviour of the MPM models to fast surrogate models in simulation. This enables us to leverage the full MPM-3DGS models for effective calibration, while increasing the simulation speed and reducing computational footprint. For soft robot actuation, we use a simplified skeleton model, and add a learned residual torque $\boldsymbol{\tau}_{r}$ to the base actuation torque of the MPM model $\boldsymbol{\tau}_\mathrm{mpm}$ to obtain the torque for the skeleton $\boldsymbol{\tau}_{s}$, where $\boldsymbol{\tau}_{s}=\boldsymbol{\tau}_\mathrm{mpm}+\boldsymbol{\tau}_{r}$. For tactile rendering, we aim to leverage an efficient probe grid simulation that captures indentation during contact as a surrogate. We then learn a mapping perception network $f_m$ between the deformation of the probe layer $\boldsymbol{\Delta}_\mathrm{probe}$ and of the surface layer of the MPM model $\boldsymbol{\Delta}_\mathrm{mpm}$, where $f_m : \boldsymbol{\Delta}_\mathrm{probe} \to \boldsymbol{\Delta}_\mathrm{mpm}$. During simulation, tactile images are then rendered from the predicted indentations.

\subsection{Tactile-guided Demonstration Collection}
\label{sec:method:demo}
To generate realistic demonstrations in the simulator, we collect a small set of real demonstrations and bootstrap a more diverse simulated demonstration set from them. Each real demonstration is a trajectory $\mathbf{y} = (\boldsymbol{\xi}_{1:T}, \mathbf{x}_{1:T}, p_{1:T},\mathbf{I}_{1:T}, \mathbf{a}_{1:T})$ of robot joint angles, end-effector positions, pressures, real tactile images, and commands.

For simulation, we focus on the contact phase in demonstrations, where the soft finger is pressurised to make contact with the manipulated object. We use the real-world demonstration trajectory and the tactile observation as an additional guidance signal for running a model-predictive controller with the cross-entropy method (CEM) in simulation~\cite{rubinstein1999cross}, using the demonstrations as warm starts with additional jittering. This allows us to collect additional demonstrations in simulation with the same observation space and action space as the real world. The tactile guidance also encourages agreement with the demonstrated contact-area trajectory to support policy training. 
\section{Experiments}
\label{sec:exps}

To demonstrate PneuTac, we design a pneumatic soft finger with a Digit vision-based tactile sensor mounted at the tip~\cite{lambeta2020digit}. The robot consists of one bellow-shaped chamber with a side-spine to enable bending. The Digit sensor is attached via a connection to the tip of the soft finger that blends with the sensor case. The soft part of the finger is 3D printed using a Stratasys printer with Agilus30 material, whereas the rigid part is printed using Vero. The gel of the Digit sensor is manufactured using Ecoflex 00-31 Near Clear. To evaluate generalisability, we additionally deploy the framework on a structurally different 3D-printed three-chamber pneumatic soft arm with similar materials, and on an unmodified Digit sensor. Runtime evaluation is performed on a single NVIDIA RTX 5070 Ti GPU. Uncertainties are estimated through the standard error of the mean (SEM).

\subsection{Real-to-Sim Modelling}
\label{sec:exps:calibration}

\subsubsection{Soft robot} 
We first perform marching cubes to generate a filled mesh from the CAD of the soft robot~\cite{lorensen1987marching}. Then, we build the MPM model in the Genesis simulator by sampling from the mesh model and using the Genesis implementation of the Neo-Hookean constitutive model. For training the 3DGS appearance model, we leverage an off-the-shelf implementation of 3DGS on a segmented image of the soft finger when it is un-actuated in a straight form using the Adam optimiser with learning rate of 0.01~\cite{kingma2015adam}. 

For calibration of the finger model parameters, we activate the chamber to 6 pre-set pressure points $P_\mathrm{train}$ both in simulation and in the real world. Before proceeding to each pressure level, the finger is first returned to resting position by setting pressure to zero to mitigate potential non-linear effects. We then render the simulated deformed finger through 3DGS. Each rendering is compared to the corresponding segmented real-world
observation using an objective that combines a masked appearance loss
with a silhouette-overlap term. For each $p\in P_{\mathrm{train}}$, let
$\mathbf{V}^{p}$ denote the observed RGB image and $\mathbf{M}^{p}$ its
binary foreground mask. Let $\hat{\mathbf{V}}^{p}$ denote the corresponding
3DGS-rendered image and $\mathbf{A}^{p}$ its binary silhouette, obtained by
thresholding the accumulated Gaussian splat opacity. Both
$\hat{\mathbf{V}}^{p}$ and $\mathbf{A}^{p}$ depend on the calibration
parameters $\boldsymbol{\theta}$, which is suppressed below
for compactness. The calibration objective averages the loss over
$P_{\mathrm{train}}=\{0,3,6,9,12,15\}\,\mathrm{kPa}$:
\begin{equation}
J(\boldsymbol{\theta})
=
\frac{1}{|P_{\mathrm{train}}|}
\sum_{p\in P_{\mathrm{train}}}
\big[
\mathcal{L}_{\mathrm{app}}
+\lambda_s(1-\operatorname{IoU})
\big].
\label{eq:calibration_objective}
\end{equation}
In each summand, $\mathcal{L}_{\mathrm{app}}$ and
$\operatorname{IoU}$ are evaluated using the corresponding
pressure-$p$ images and masks:
\begin{equation}
\mathcal{L}_{\mathrm{app}}
\big(\mathbf{M}^{p},\hat{\mathbf{V}}^{p},\mathbf{V}^{p}\big)
=
\frac{
\displaystyle
\sum_{i,j} M_{ij}^{p}
\left[
\frac{1}{3}\sum_{k=1}^{3}
\left|\hat{V}_{ijk}^{p}-V_{ijk}^{p}\right|
\right]
}{
\displaystyle
\sum_{i,j} M_{ij}^{p}
},
\label{eq:masked_photometric}
\end{equation}
where $i,j$ index pixel rows and columns, and $k$ indexes the RGB
channels. The silhouette intersection-over-union is
\begin{equation}
\operatorname{IoU}(\mathbf{M}^{p},\mathbf{A}^{p})
=
\frac{
\displaystyle
\sum_{i,j} M_{ij}^{p}A_{ij}^{p}
}{
\displaystyle
\sum_{i,j}
\left(M_{ij}^{p}+A_{ij}^{p}-M_{ij}^{p}A_{ij}^{p}\right)
},
\label{eq:silhouette_iou}
\end{equation}
so that $1-\operatorname{IoU}$ penalises silhouette mismatch.
For RGB intensities normalised to $[0,1]$ and nonempty foreground
masks, both $\mathcal{L}_{\mathrm{app}}$ and $\operatorname{IoU}$
lie in $[0,1]$. We set $\lambda_s=2$.

The system identification focuses on the material parameters $\boldsymbol{\theta}_m$, including Young's modulus $E_s$ and Poisson's ratio $\nu_s$, and the actuation parameters $\boldsymbol{\theta}_a$, which include the pressure-to-bending gain $G$, the per-joint gain-distribution parameters $\alpha$ and $\beta$, and a residual rest-pressure $p_0$ for the multi-joint spine. Concretely, $N_\mathrm{finger}{=}4$ spine joints are defined, ordered $n=1$ at the base to
$n=N_\mathrm{finger}$ at the tip, and driven to $q^{cmd}_n = G\,r_n(\alpha,\beta)\,(p + p_0)$, with
per-joint profile
\begin{equation}
r_n(\alpha,\beta) \;\propto\; \exp\!\big(\alpha\,(n-\bar n)\big)\,
\big(1 + \beta\,(n-\bar n)^2\big),
\;\;
\bar n = \tfrac{N_{\text{finger}}+1}{2}
\label{eq:gain-profile}
\end{equation}
normalised to unit mean, $\tfrac{1}{N_\mathrm{finger}}\sum_n r_n = 1$. The gain $G$ therefore
sets the overall bending per unit pressure, while $\alpha$ and $\beta$
redistribute it along the spine. The offset $p_0$ is a residual rest pressure capturing the bend already
present when the chamber is un-actuated. The target spine joint angle $q^{cmd}$ is then fed to a controller to compute the torque $\tau_{mpm}$. We jointly optimise the $\boldsymbol{\theta}_m$ and $\boldsymbol{\theta}_a$ parameters using a derivative-free evolutionary strategy CMA-ES (population 7, 84 evaluations, initial step 15\% of each gain, warm-started from a coarse fit)~\cite{hansen2001complete}. We use a marker for ground-truth motion of the tip, and leverage held-out pressure levels for evaluation: $P_{\mathrm{eval}} = P_{\mathrm{all}} \setminus P_{\mathrm{train}}$,
where $P_{\mathrm{all}} = \{0,1,\ldots,15\}\,\mathrm{kPa}$.

For baseline comparison, we implement three alternative rendering methods, all using the same MPM particle geometry, and compare the root-mean-square error (RMSE): 1) Textured mesh, obtained by extracting a surface with marching cubes and assigning each vertex the colour of its nearest particle~\cite{lorensen1987marching}; 2) Particle splat, which renders each particle as a fixed-radius coloured disk; 3) Deformable NeRF, a canonical-space neural radiance field rendered at each pressure by warping every ray sample into canonical space through the MPM particle displacement field~\cite{mildenhall2021nerf}.

For the three-chamber soft arm, we leverage the same modelling procedure with a six-joint ($N_{arm}{=}6$) formulation, and apply the training pressure levels \(P_{\mathrm{train}}\) to each chamber separately. We additionally include an orthogonal observation of the soft arm from the side to capture the out-of-plane bending relative to the primary view.

\subsubsection{Tactile rendering} 
For the tactile calibration, we first generate the gel's MPM model by regular grid sampling of particles within a box matching the gel's dimensions, leveraging known values of the material parameters of the gel. Next, using the known Digit camera parameters, we train the coupled 3DGS model on a single background image $\mathbf{I}_{rest}$, using Adam with learning rate of 0.01. 

For the contact photometric model, we press an indenter on the real sensor to obtain one indentation image $\mathbf{I}_{indent}$, and simulate the corresponding MPM deformation $\boldsymbol{\Delta}_{mpm}$ with the same indenter. From this pair we fit the coefficients $\{\boldsymbol{\ell}_x,\boldsymbol{\ell}_y,\mathbf{c}_d\}$, which are identified in two steps. First, we perform a sinusoidal regression of the per-pixel colour residual $\mathbf{I}_{indent} - \mathbf{I}_{rest}$ against the local slope angle $\text{atan2}(g_y,g_x)$, which yields $\boldsymbol{\ell}_x, \boldsymbol{\ell}_y$. Meanwhile, a weighted least-squares fit of the post-tilt residual against the normalised indentation depth $\bar{d}_g$ yields $\mathbf{c}_d$. Second, the values are refined through the renderer by gradient-based optimisation using a loss of $\lVert\hat{\mathbf{I}} - \mathbf{I}_{indent}\rVert_2^2$ with Adam at a learning rate of $5\times10^{-3}$. The same procedure is repeated on both the sensor attached to the soft finger and the unmodified Digit sensor.

For evaluation, we use faithful reimplementations of the Taxim, Tacchi, and DOT-Sim models ~\cite{si2022taxim,chen2023tacchi,sun2024tacchi,you2026dot} and provide the same training and evaluation datasets. We compare the rendered tactile images on a held-out set of 20 real tactile images of various indentations, using RMSE, peak signal-to-noise ratio (PSNR), and normalised-cross-correlation (NCC) metrics. 

\subsection{Action and Perception Networks}
\label{sec:exps:networks}
The action network and the perception network are trained from data collected in simulation. For the action network, we collect joint trajectories ${q, \dot q}$ of the multi-joint spine of the full MPM model under varied pressure inputs $p$. We use an MLP network to learn $f_r(\mathbf{q}, \dot {\mathbf{q}}, p)$, which outputs the residual joint torque $\boldsymbol{\tau}_r$, so the total applied torque on the skeleton model is $\boldsymbol{\tau}_s = \boldsymbol{\tau}_{mpm}(p) + \boldsymbol{\tau}_r$. The network is trained by minimising the mean-squared error (MSE) between its prediction and the analytic target $\boldsymbol{\tau}_r^\star = \mathbf{M}_{\mathrm{skel}}\ddot{\mathbf{q}} + \mathbf{D}\dot{\mathbf{q}} + \mathbf{K}\mathbf{q} - \boldsymbol{\tau}_{\mathrm{mpm}}(p)$, obtained by inverse dynamics from the full-model trajectories in simulation. We collect 8,000 state–torque pairs for training. The network is a fully connected MLP with two hidden layers of width $128$. For training, we use Adam at a learning rate of $1\times10^{-3}$.
    
For the perception network, we collect indentation depth maps captured by the probe and the MPM model $\{\boldsymbol{\Delta}_\mathrm{probe},\boldsymbol{\Delta}_\mathrm{mpm}\}$ during paired contact simulation. We collect 100 samples using varied indentation shapes, depths, rotations and positions in simulation, and train a UNet to map between the depth observations~\cite{ronneberger2015unet}. The simulated probe deformation grid ${\boldsymbol{\Delta}}_\mathrm{probe}$ is constructed to the same dimension as the MPM-equivalent field $\boldsymbol{\Delta}_\mathrm{mpm}$. The UNet perception mapping network is a three-level convolutional encoder-decoder with ReLU and skip connections that maps $\boldsymbol{\Delta}_\mathrm{probe}$ to $\boldsymbol{\Delta}_\mathrm{mpm}$. We train by minimising MSE over the contact region with the Adam optimiser at a learning rate of $1\times10^{-3}$.

\subsection{Contact-Rich Compliant Manipulation}
\label{sec:exps:demonstration}
To demonstrate the effectiveness of the PneuTac framework, we leverage a set of contact-rich soft robot manipulation tasks in a policy learning pipeline, where the soft finger is attached to the end-effector of a Panda arm. Real-world demonstrations are then collected manually for three delicate manipulation tasks that are challenging for rigid robots to perform safely, which include:
\begin{enumerate}
\item Switch flipping: flip a toggle switch.
\item Egg carton opening: find contact with the edge of the carton to open the carton lid.
\item Card pulling: pull out a single poker card from a deck standing in a container.
\end{enumerate}
We collect 10 demonstrations for each task, and use these to bootstrap 100 additional demonstrations each in simulation. 

\subsubsection{Demonstration collection in simulation} 
Rather than using a single endpoint descriptor as a sparse MPC objective, we extract the full demonstrated trajectory with tactile feedback for guidance. Demonstrations are collected in simulation by a batched, GPU-parallel
CEM-MPC planner, warm-started from human teleoperation demonstrations.
For each real-world demonstration, we plan over a horizon $H$, and the planner minimises the
trajectory-tracking cost of~\cref{eq:mpc-impl}, where different scales account for the different units:
\begin{equation}
\mathcal{L}_\mathrm{dem}=\sum_{t=1}^{H} \Big( w\,\big(\hat c_t - c^\star_t\big)^2
   + \lambda\,\big\lVert \hat{\mathbf{x}_t} - \mathbf{x}^\star_t \big\rVert^2 \Big)
   \;+\; \mu \sum_{t=2}^{H} \big\lVert a_t - a_{t-1} \big\rVert^2
\label{eq:mpc-impl}
\end{equation}
where $c_t = s_t/N$ denotes the contact-area readout of the tactile signal, $s_t$ is the number of contact pixels obtained by thresholding the difference between each tactile image $\mathbf{I}_t$ and a background reference $\mathbf{I}_\mathrm{rest}$ at step $t$, $N$ is the total number of pixels, and $a_t$ denotes the action at step $t$. The weights are shared across tasks: $w{=}1.0$, $\lambda{=}50$, $\mu{=}0.01$. The first term tracks the demonstrated tactile-property trajectory, the second anchors the trajectory to the demonstrated success trajectory, and the third penalises jerky actions.

Collection is run with per-variant domain randomisation, perturbing four factors: (i) a friction
feed-forward term; (ii) a coupled object-position shift, sampled
$\mathcal{N}(0,\sigma^2)$ per axis with $\sigma$ set to the measured
real-demonstration contact-onset position standard deviation; (iii) a peak-pressure jitter; and (iv) a pressure-to-bend gain
multiplier for the soft finger. 

\subsubsection{Policy training} 
For policy learning, we use a simple behaviour cloning setup~\cite{bain1999framework}. The policy outputs an end-effector Cartesian displacement and a finger pneumatic pressure, $\mathbf{a} = \{\Delta\mathbf{x}, p\}$ with $\Delta\mathbf{x} \in \mathbb{R}^3$, where orientation is held fixed. Policies are trained with Adam using a learning rate of $3\times10^{-4}$. The
loss is a sum of the squared end-effector error
and the squared pressure error,
$\mathcal{L}_\mathrm{policy}=\overline{\lVert 1000\,(\Delta \mathbf{\hat x}-\Delta \mathbf{x}^\star)\rVert^2}
 +\overline{(\hat p-p^\star)^2}$, where the scale is due to the different units in the displacement and pressure terms.
 
The BC policy consumes a $24$-D observation of joint states, end-effector positions, contact area, pressure, and the previous action, where $\mathbf{o}_t=\{\boldsymbol{\xi}_t, \boldsymbol{\dot \xi}_t, \mathbf{x}_t,s_t/N, \log(1+s_t), p_t,\mathbf{a}_{t-1}\}$, as well as a tactile image. As the soft pneumatic finger only exerts a limited amount of force and is only able to create a small amount of indentation, we use both the normalised contact area $s_t/N$ and log-compressed contact-pixel count $\log(1 + s_t)$ as additional scalar tactile inputs.

We compute the absolute difference between the tactile image and its background reference, which is then encoded by a convolutional encoder with a linear layer to a $64$-D embedding. This embedding is concatenated
with the $24$-D scalar observation and passed through an MLP with two 256-unit hidden layers and a 4-D action output. 

We train $\pi_\mathrm{Sim+Real}$ with demonstrations collected in simulation, and fine-tune with the real-world demonstrations. The same set of real-world demonstrations is used for both the bootstrapping and the policy learning steps. For evaluation, we compare the following training conditions: 1) Real: BC policy trained with the same set of 10 real-world demonstrations only; 2) N-Aug: naive augmentation without the simulator using time warping, demonstration mix-up, kinematic/pressure jitter, and tactile appearance jitter. It generates 100 augmented trajectories from the same 10 real demonstrations and uses the same pre-training and real-data fine-tuning protocol as Sim+Real; 3) Real-20: BC policy trained on 20 real demonstrations, without simulation pre-training; 4) No-Tac: the full sim+real training pipeline with the tactile observation removed; 5) PPO with fine-tuning~\cite{schulman2017proximal}. We train an asymmetric actor–critic PPO in PneuTac with BC's action space, where the actor observes proprioception and the probe grid, and the critic additionally observes the object state. The reward sums progress, approach, contact, tactile-depth, success, and action-rate terms. For deployment, the actor is distilled into a student with the same observation space as the BC policy, and given the identical sim-to-real fine-tuning.
\section{Results and Discussion}
\label{sec:results}

We evaluate PneuTac by the quality of the real-to-sim modelling, the efficiency of the surrogate models, and the value of tactile-guided simulation demonstrations for policy learning on compliant manipulation tasks.

\subsection{Real-to-Sim Modelling}

By using the CAD models available for the soft robot and tactile sensor, we sample $\sim$4k particles for modelling the soft robot, and $\sim$60k particles for modelling the soft gel. The higher resolution for the soft gel is needed for accurately modelling its deformation under contact. 

\subsubsection{Soft robot finger}

\begin{figure}[ht]
    \centering
    \includegraphics[width=0.75\columnwidth]{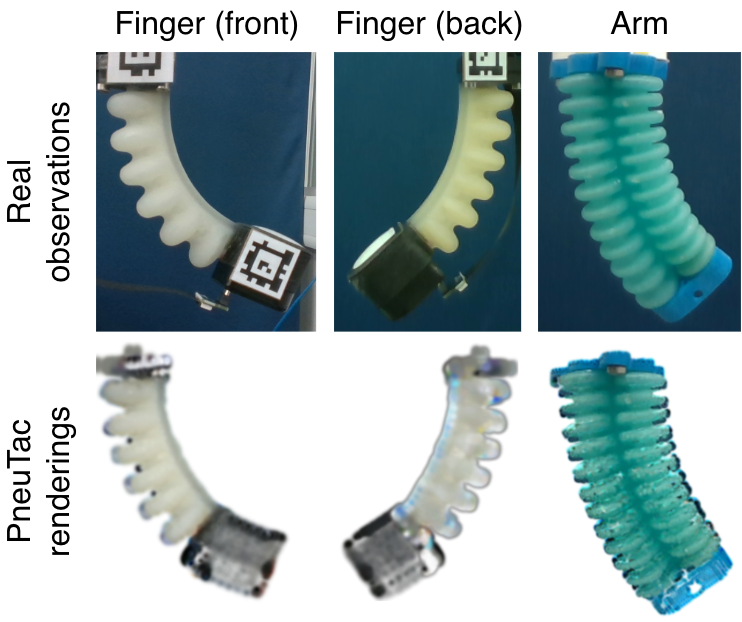}
    \caption{Real-to-sim comparison of the soft finger and three-chamber arm in novel configurations. The finger's back view is rendered from the side opposite the calibration camera.}
    \label{fig:result:softrobot}
\vspace{-3mm}
\end{figure}

By leveraging the MPM model of the soft robot with a rigid spine for motion, PneuTac is able to accurately simulate the behaviour of the soft finger. From the vision-based calibration procedure, we identify the parameter values as $G=0.01114$~rad/kPa,
$\alpha=0.0313$, $\beta=0.2974$, $p_0=10.70$~kPa, $E_s=0.945$~MPa and $\nu_s=0.460$. These values are
physically realistic, where \(E_s\) sits in the regime expected from the printer, and the actuation gain
\(G\) with per-joint profile \(r_n(\alpha,\beta)\) also reproduces the tip deflection accurately,
while $p_0$ absorbs the gravity-induced rest curl when the robot is un-actuated. As seen in \cref{fig:result:softrobot}, the PneuTac simulation is able to reproduce the bending shape accurately for both the soft finger and the soft arm using the coupled MPM-3DGS representation. We additionally evaluate the soft finger appearance model by rendering from the back side of the robot, opposite the calibration camera. Even at the novel viewpoint, the rendered observation in \cref{fig:result:softrobot} matches the corresponding real observation. As seen in \cref{table:results:body}, compared to baseline methods, the PneuTac framework is able to render both front and back views of the soft finger realistically, demonstrating the performance of the framework and its viewpoint generalisability. 

\begin{table}[ht]
\centering
\caption{RMSE comparison of soft finger rendering methods for front and back
views. Values are reported as mean $\pm$ SEM.}
\label{table:results:body}
\setlength{\tabcolsep}{2.5pt}

\begin{tabular}{@{}lcccc@{}}
\toprule
\textbf{View} &
\textbf{PneuTac} &
\textbf{Tex-Mesh} &
\textbf{Par-Splat} &
\textbf{Def-NeRF} \\
\midrule

Front $\downarrow$ &
$\mathbf{50.07 \pm 0.83}$ &
$51.13 \pm 1.12$ &
$54.58 \pm 1.59$ &
$55.06 \pm 11.72$ \\

Back $\downarrow$ &
$77.61 \pm 11.72$ &
$79.09 \pm 9.08$ &
$78.62 \pm 8.43$ &
$\mathbf{68.96 \pm 12.77}$ \\

\bottomrule
\end{tabular}
\end{table}

We also quantitatively evaluate the predicted motions of the simulated soft robots with marker-based real-world tip measurements at pressure points that are \emph{unseen} during the parameter optimisation. Compared to the marker ground-truths, the model achieves a tip tracking error of 4.2 $\pm$ 0.4~mm, less than 4\% of the finger length, and 4.9 $\pm$ 1.0~mm, 4.2\% of the arm length. Both results are also commensurate with prior approaches on similar platforms~\cite{zhong2025end,dubied2022sors} yet PneuTac requires only a simple camera setup.

\subsubsection{Tactile rendering}
After construction of the gel model, we obtain a $59\times80\times13$ grid of MPM particles coupled to a 3DGS appearance model, and use that to generate realistic tactile observations. Compared to baseline models, the framework outperforms in all evaluated image metrics given the same training dataset, shown in \cref{table:results:tactile}, demonstrating that PneuTac is able to generate more realistic tactile image observations. This illustrates the advantage of the framework in the small calibration dataset regime. When evaluated on another unmodified Digit sensor, the framework achieves RMSE 8.12 $\pm$ 0.22, PSNR 29.97 $\pm$ 0.22 dB, and NCC 0.976 $\pm$ 0.001 on evaluation images, on par with the original sensor. Directly reusing the original calibration gives RMSE 22.96, indicating the device differences. These results show that the PneuTac framework is able to generalise beyond the specific hardware used in the main experiments.

\begin{table}[t]
\centering
\caption{Comparison of tactile rendering methods. Values are reported as mean $\pm$ SEM.}
\label{table:results:tactile}
\footnotesize
\setlength{\tabcolsep}{2.5pt}

\begin{tabular}{@{}lcccc@{}}
\toprule
\textbf{Metric} &
\textbf{PneuTac} &
\textbf{Tacchi} &
\textbf{DOT-Sim} &
\textbf{Taxim} \\
\midrule

RMSE $\downarrow$ &
$\mathbf{10.18 \pm 0.58}$ &
$10.86 \pm 0.29$ &
$14.14 \pm 0.23$ &
$14.13 \pm 0.61$ \\

PSNR (dB) $\uparrow$ &
$\mathbf{28.11 \pm 0.54}$ &
$27.44 \pm 0.23$ &
$25.13 \pm 0.14$ &
$25.20 \pm 0.36$ \\

NCC $\uparrow$ &
$\mathbf{0.982 \pm 0.001}$ &
$0.952 \pm 0.004$ &
$0.913 \pm 0.004$ &
$0.915 \pm 0.011$ \\

\bottomrule
\end{tabular}
\vspace{-3mm}
\end{table}

\begin{figure}[h]
    \centering
    \includegraphics[width=\columnwidth]{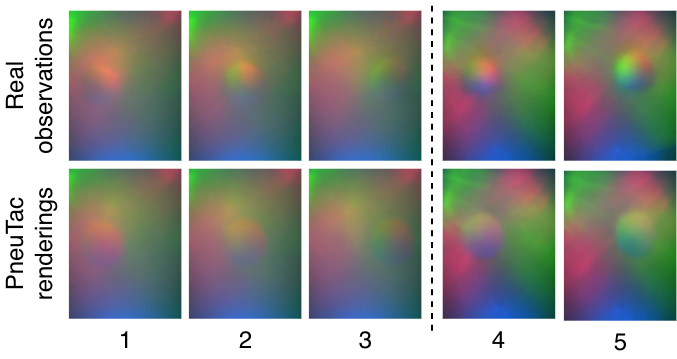}
    \caption{Tactile image comparison. Columns 1-3: the Digit sensor attached to the soft finger; columns 4-5: a second, unmodified Digit sensor.}
    \label{fig:rendering}
    \vspace{-3mm}
\end{figure}

Qualitatively, as seen in~\cref{fig:rendering}, the framework is able to capture the lighting change from more red regions (columns 1 and 4), to the more green portions (columns 3 and 5). The results demonstrate the utility of the contact photometric model in the PneuTac framework in capturing the lighting change due to contact and for efficient real-to-sim modelling of the vision-based tactile sensors.

\subsection{Action and Perception Networks}
We first evaluate the MLP residual torque network $f_r$, whose output $\tau_r$ is added to base actuation torques to obtain the skeleton torque $\tau_s$ to match the behaviour of the full MPM model. After the residual torques are applied on the skeleton model, the model achieves a mean joint position error of 0.051 $\pm$ 0.006° and velocity error of 2.76 $\pm$ 0.55°/s compared to the full MPM simulation. Meanwhile, using the skeleton model achieves a 5x speedup in comparison (\raisebox{-0.65ex}{\textasciitilde}10ms for action network vs \raisebox{-0.65ex}{\textasciitilde}50ms for full MPM), enabling efficient simulation of the soft finger.

We then evaluate the perception network. Using a UNet to map the indentation depth reported by the grid probe sensors to that from the MPM simulation, we are able to achieve a mean absolute error of 0.077 $\pm$ 0.005 mm for each contact point, out of a mean indentation depth of 1.4 mm. This demonstrates that the perception network is able to accurately map the indentation depth observation from the fast probe model to the MPM model. Additionally, using the probe model achieves a 3000x speedup in simulation speed (\raisebox{-0.65ex}{\textasciitilde}50ms vs \raisebox{-0.65ex}{\textasciitilde}150s), due to the large number of particles in the high-fidelity MPM simulation. This enables us to efficiently and accurately simulate the tactile indentation for policy learning. 

\subsection{Contact-Rich Compliant Manipulation}

\begin{figure}[ht]
    \centering
    \includegraphics[width=0.31\columnwidth]{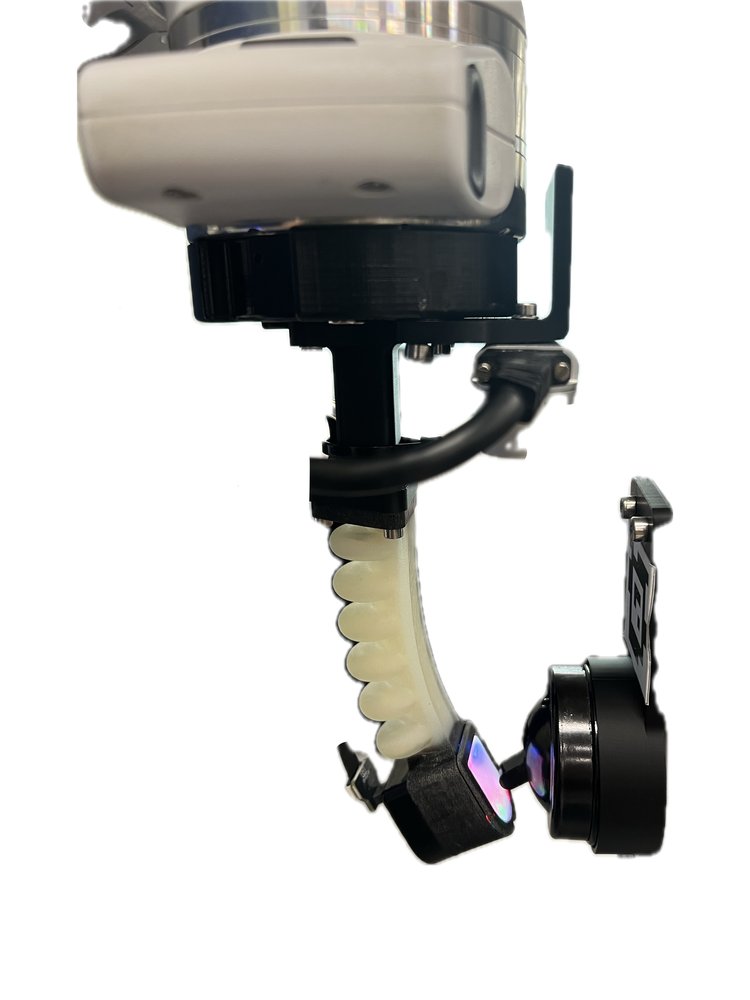}
    \hfill
    \includegraphics[width=0.31\columnwidth]{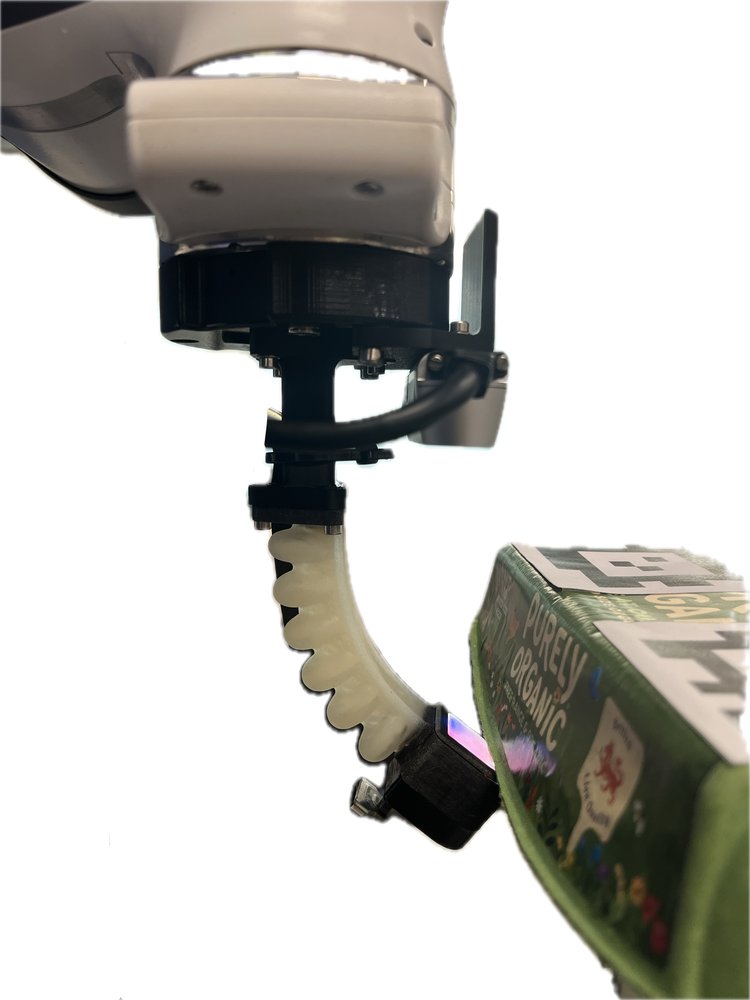}
    \hfill
    \includegraphics[width=0.31\columnwidth]{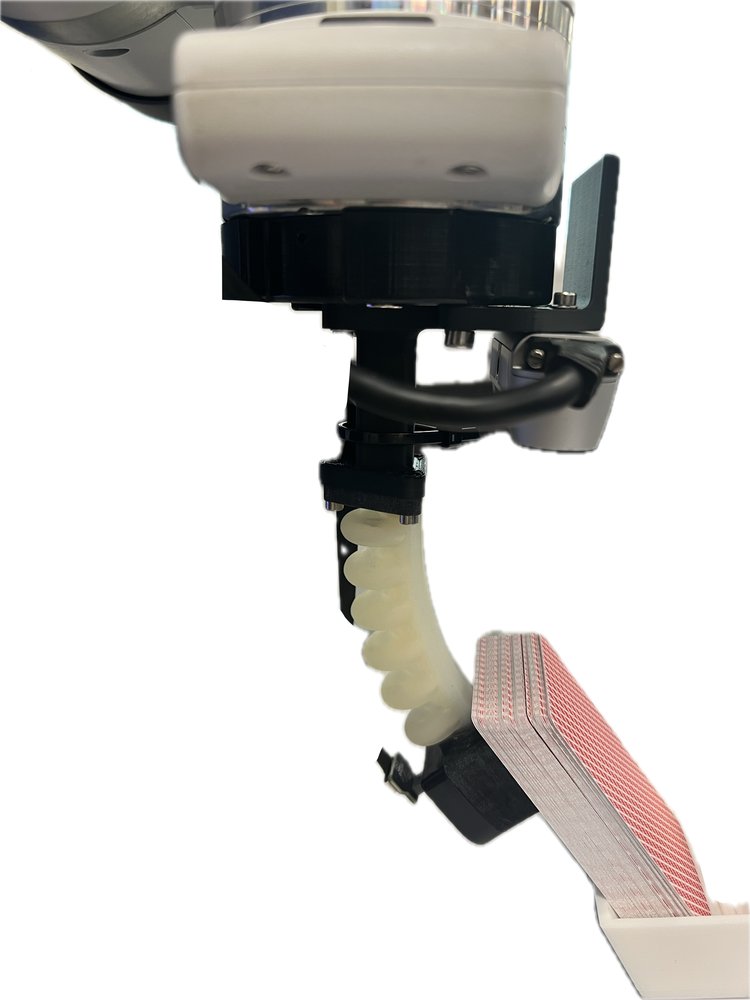}
    \caption{Contact-rich manipulation with the tactile soft finger. Left to right: switch flipping, egg carton opening, and card pulling.}
    \label{fig:tasks}
\vspace{-3mm}
\end{figure}

We evaluate the results of the policy trained using the simulation dataset with real-world fine-tuning $\pi_\mathrm{Sim+Real}$ and the baseline policies in three contact-rich manipulation tasks shown in~\cref{fig:tasks}. In each task, the policy needs to pressurise the pneumatic chamber and control the robot to make contact with the object to facilitate manipulation. Each task is evaluated for 10 trials per seed across 3 seeds. The results of the policy deployment are shown in ~\cref{table:results:policy}.

\begin{table}[t]
\centering
\caption{Real-world task success rates (\%). Values are reported as mean $\pm$ SEM across 3 seeds, with 10 trials per seed and task.}
\label{table:results:policy}

\setlength{\tabcolsep}{1.25pt}

\begin{tabular}{@{}lcccccc@{}}
\toprule
\textbf{Task} &
\textbf{Sim+Real} &
\textbf{Real} &
\textbf{N-Aug} &
\textbf{Real-20} &
\textbf{No-Tac} &
\textbf{PPO} \\
\midrule

Switch
& $83.3 \pm 3.3$
& $60.0 \pm 5.8$
& $73.3 \pm 3.3$
& $\mathbf{90.0 \pm 5.8}$
& $30.0 \pm 5.8$
& $63.3 \pm 3.3$ \\

Egg
& $\mathbf{90.0 \pm 5.8}$
& $26.7 \pm 8.8$
& $46.7 \pm 6.7$
& $40.0 \pm 10.0$
& $56.7 \pm 3.3$
& $80.0 \pm 5.8$ \\

Card
& $\mathbf{66.7 \pm 3.3}$
& $33.3 \pm 8.8$
& $40.0 \pm 5.8$
& $50.0 \pm 5.8$
& $63.3 \pm 3.3$
& $10.0 \pm 5.8$ \\

\bottomrule
\end{tabular}
\vspace{-3mm}
\end{table}

As seen, leveraging the additional demonstrations collected from the simulation, we are able to substantially increase the policy success rates when deployed in the real world, compared to using the same amount of real-world demonstrations for training. This demonstrates the downstream utility of PneuTac in accurately modelling soft pneumatic robots with tactile sensors. The naive augmentation results show that generic augmentation improves over the real-only baseline but does not match Sim+Real, and the tactile- and trajectory-guided demonstrations from PneuTac provide useful information for policy pre-training. Comparing against Real-20, for the egg carton and card tasks, the Sim+Real policy performs considerably better, indicating that simulation-augmented demonstrations are effective where real data potentially under-samples the success manifold. On the switch task, where 10 real demonstrations are already informative, doubling the real budget matches the simulation-augmented pipeline. The no-tactile result shows that tactile feedback substantially improves success on switch and egg carton opening, but the observed difference on card pulling is small, consistent with card pulling being position- rather than contact-dominated. The PPO baseline is comparable to the BC pipeline on the egg carton task and trails it on the switch task. Its failure on card is consistent with the simulator's reduced fidelity for card-on-card contact, to which PPO's reward-driven exploration is likely more exposed than the BC policy anchored on real demonstrations.

These experiments demonstrate the utility of PneuTac-augmented demonstrations for tactile manipulation on the soft finger platform. A direct policy baseline comparison with another simulator is not available because, to the best of our knowledge, few existing simulation frameworks are capable of natively modelling the paired devices. PneuTac is among the first frameworks that jointly model these devices within a policy-learning pipeline, and opens the way to tactile-based compliant manipulation with soft robots. 
\section{Conclusion}
\label{sec:Conclusion}
In conclusion, we propose PneuTac, a unified framework for tactile-feedback manipulation using soft pneumatic robots, based on MPM-3DGS simulation. The framework enables efficient real-to-sim modelling with a simple vision-based method, validated on different soft robot and sensor devices, and outperforming prior baselines in realism. Using action and perception networks, the framework also achieves efficient simulation of the challenging behaviours of the soft robot finger and the tactile sensor. Additionally, we demonstrate that the framework is also useful for tactile-based policy learning in a demonstration-augmentation pipeline, through a set of real-world contact-rich compliant manipulation tasks. This makes PneuTac a practical engine for tactile manipulation on compliant hardware.


\bibliographystyle{IEEEtran}
\bibliography{references}

\end{document}